\documentclass{article}

\usepackage{microtype}
\usepackage{graphicx}
\usepackage{subcaption}
\usepackage{booktabs} 

\usepackage{hyperref}

\newcommand{\vectors}[1]{\lowercase{\mathbf{#1}}}

\usepackage{algorithm}
\usepackage{amsmath}
\usepackage{amsfonts}
\usepackage{enumitem}

\usepackage[accepted]{icml2018}

\icmltitlerunning{Graph Node-Feature Convolution for Representation Learning}

\begin{document}

\twocolumn[
\icmltitle{Graph Node-Feature Convolution for Representation Learning}
\icmlsetsymbol{equal}{*}
\begin{icmlauthorlist}
\icmlauthor{Li Zhang}{Sheffield}
\icmlauthor{Heda Song}{Nottingham}
\icmlauthor{Nikolaos Aletras}{Sheffield}
\icmlauthor{Haiping Lu}{Sheffield}
\end{icmlauthorlist}
\icmlaffiliation{Sheffield}{Department of Computer Science, University of Sheffield, Sheffield, UK}
\icmlaffiliation{Nottingham}{Department of Computer Science, University of Nottingham, Nottingham, UK}
\icmlcorrespondingauthor{Li Zhang}{lzhang72@sheffield.ac.uk}
\icmlcorrespondingauthor{Heda Song}{Heda.Song@nottingham.ac.uk}
\icmlcorrespondingauthor{Nikolaos Aletras}{n.aletras@sheffield.ac.uk}
\icmlcorrespondingauthor{Haiping Lu}{h.lu@sheffield.ac.uk}

\icmlkeywords{Machine Learning, ICML}

\vskip 0.3in
]

\printAffiliationsAndNotice{} 

\begin{abstract}
Graph convolutional network (GCN) is an emerging neural network approach. It learns new representation of a node by aggregating feature vectors of all neighbors in the aggregation process without considering whether the neighbors or features are useful or not. Recent methods have improved solutions by sampling a fixed size set of neighbors, or assigning different weights to different neighbors in the aggregation process, but features within a feature vector are still treated equally in the aggregation process. In this paper, we introduce a new convolution operation on regular size feature maps constructed from features of a fixed node bandwidth via sampling to get the first-level node representation, which is then passed to a standard GCN to learn the second-level node representation. Experiments show that our method outperforms competing methods in semi-supervised node classification tasks. Furthermore, our method opens new doors for exploring new GCN architectures, particularly deeper GCN models. 

\end{abstract}

\section{Introduction}
\begin{figure*}
	\includegraphics[height=3in, width=7in]{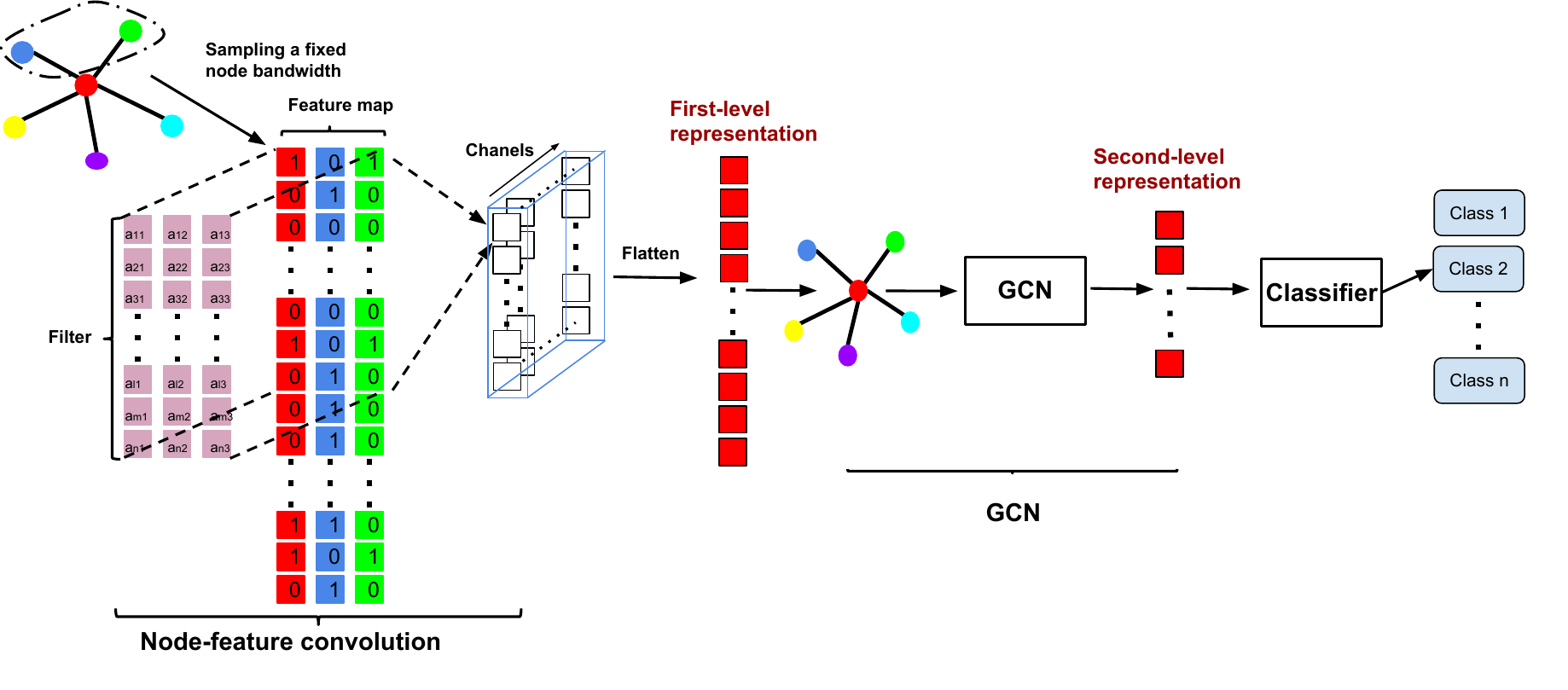}
	\caption{An illustration of the proposed graph convolutional networks with node-feature convolution (NFC-GCN) for graph representation learning, with three main steps. 1) \textit{Neighbor sampling}: randomly sample neighbors for each node with a fixed node bandwidth. 2) \textit{Node-feature convolution:} perform convolution operation on 2D node-feature maps and flatten the obtained vectors to get the first-level node \textit{Representation 1}. 3) \textit{Standard GCN:} Feed the first-level NFC representation to a standard GCN to learn a second-level node \textit{Representation 2}. After representation learning steps, send the second-level representation to the classifier to predict the node class label. (This figure is best viewed in color / on screen).}
	\label{tab:newmodel}
\end{figure*} 

Graphs, such as social networks, biological networks, and citation networks, are ubiquitous data structures that can capture interactions between individual nodes \cite{Hamilton2017RepresentationLO}. Nodes in graphs are often associated with feature vectors. For example, in a typical citation graph, nodes are documents, edges are citation links, and node features are bag-of-words feature vectors. This paper will focus on analyzing such graphs with node features available.

Graphs are challenging to deal with \cite{Shaw2009StructurePE}. Most real-world graphs have no regular connectivity and the node degrees can range from one to hundreds or even thousands in the same graph. Moreover, graphs have rich and important information that can not be revealed by simply analyzing the individual nodes or structure information. For example, sparse bag-of-words feature vectors can not effectively reflect the citations between papers. To understand complex graphs better, it is important to learn graph representations that can capture rich information from both node feature vectors and graph structures (i.e., node neighborhood information). 

\citet{kipf2016semi} proposed graph convolutional networks (GCN) as an effective graph representation model that naturally combines structure information and node features in the learning process. It represents a node by aggregating all the feature vectors of its neighbors, analogous to the receptive field of a convolutional kernel in convolutional neural networks (CNN). It has been proved to be powerful in many applications, including node classification, link prediction, and recommendation \cite{kipf2016semi, schlichtkrull2018modeling, Ying2018GraphCN}. However, GCN aggregates all neighbors and does not consider whether the central node has a dense or sparse connection, and whether a neighbor's individual features are useful or not in the aggregation process. 

There are two existing approaches to address the two problems of GCN above. The first is sampling-based approach. Instead of considering all neighbors, this approach samples a fixed-size set of neighbors so that the neighborhood resembles that in CNN better.  \citet{Hamilton2017RepresentationLO} proposed GraphSAGE that randomly samples a fixed-size set of neighbors by random walk. FastGCN by \citet{Chen2018FastGCNFL} and jumping knowledge networks (JK-networks) by \citet{pmlr-v80-xu18c} sample nodes from the whole graph rather than the neighborhood, aiming to improve the efficiency. However, these methods did not consider to weight the selected node or features in the aggregation process either .

The second approach focuses on how to learn to weight neighbor feature vectors, instead of simply aggregating them. Inspired by the attention mechanism \cite{Bahdanau2014NeuralMT}, \citet{velickovic2017graph} proposed the graph attention networks (GATs) to use a 1D convolutional layer to learn different weights of neighbors for aggregation. However, in GAT, each feature of a feature vector shares the same weight, i.e., the usefulness of features is not considered and useful features are weighted the same as less useful features. Recently, \citet{Gao:2018:LLG:3219819.3219947} proposed a learnable graph convolutional networks (LGCNs). It applys a learnable graph convolutional layer after a graph embedding layer (GCN layer), then uses two 1D convolutional layers to perform convolution on features from the GCN layer. The feature map consists of the central node's embedding and reorganized neighbors' embedding, rather than the original features. Since the learnable convolutional layer operates after the GCN layer, it inherits the limitations of GCN above.

As a powerful representation learning method, CNN can successfully work on fixed-size grids (e.g., images) or sequences (e.g., sentence) datasets to tackle problems such as image classification \cite{krizhevsky2012imagenet, karpathy2014large}, semantic segmentation \cite{girshick2014rich} and machine translation \cite{Bahdanau2014NeuralMT}. GCN and its extensions above all aim to apply CNN-like convolutional operations on graphs. They have made progress in performing convolutional operation on node representations, using neighborhood for node representation (to imitate the receptive field). However, this is different from the convolution in CNN where the convolution operation works on features, and GCN and its extensions just use the connectivity structure of the graphs as the receptive field to perform neighborhood aggregation.

In this paper, we propose a novel GCN extension named as graph convolutional networks with node-feature convolution (NFC-GCN), aiming to generalize the concepts in CNN further to graphs. In particular, NFC-GCN uses the 1D or 2D convolution on node-feature 2D feature map to learn new representation of the central node. NFC-GCN has three steps of operation as shown in Fig.~\ref{tab:newmodel}.
\begin{enumerate}
	\item \textbf{Neighbor sampling:} We randomly sample a fixed-size set of neighbors for each central node so that each node has a regular connectivity. 
	\item \textbf{Node-feature convolution (NFC):} After sampling the neighbors, we propose a node-feature convolutional layer to learn different weights for a node-feature 2D feature map to get the first-level node representation via convolution and flattening. 
	\item \textbf{Standard GCN:} We feed the learned first-level representation to a standard GCN to learn a second-level node representation. 
\end{enumerate}
Therefore, the proposed NFC-GCN embodies ideas from the sample-based GCN methods \cite{Hamilton2017RepresentationLO, Chen2018FastGCNFL, pmlr-v80-xu18c} and extends the ideas of 1D convolutional layer in GAT \cite{velickovic2017graph}, while also keeping the virtues of the original, standard GCN \cite{kipf2016semi}. 

Our key idea is in the node-feature convolution step that introduces a convolutional layer to work on a 2D feature map constructed \textbf{directly} from feature vectors of the central node and its sampled neighbors. This layer enables \textit{end-to-end learning} of weights for different features from different neighbors. To reduce the model complexity (i.e., the number of model parameters) and reduce overfitting, we keep the filter size small, use multiple filters, and share a filter's parameters on all nodes in the graph. This makes the convolution in this NFC layer resembles more the convolution of CNN on images than the convolution in other GCN methods. 

Experiments on three citation graph datasets show that NFC-GCN outperforms existing GCN methods across all three datasets. In addition, it can converge with less training epochs. Furthermore, we studied deeper models of NFC-GCN and GCN with up to five GCN layers. The results show that NFC-GCN has much smaller performance variation than GCN. This encourages the exploration of deep learning models for graphs, an area with little progress so far. 

\section{Related Work}
In this section, we review and discuss related works on graph representation learning and particularly GCN and its GCN extensions.

\subsection{Notations}
In this paper, we consider graphs with a feature vector associated with each node. Let $\mathcal{G}$= $(\mathcal{V},\mathcal{E},\mathbf{X})$ denotes an undirected graph with $N$ nodes $v_i \in \mathcal{V}$, edges $(v_i, v_j)\in\mathcal{E}$, where $i,j=1,\cdots N$, an adjacency matrix $\mathbf{A}\in\mathbb{R}^{N\times N}$, and a feature matrix $\mathbf{X} \in \mathbb{R}^{N \times D}$ containing the $N$ $D$-dimensional feature vectors. We first define a list of important notations that will be used throughout this paper, as shown in Table~\ref{tab:notation}.
\begin{table}[tbp]
	\renewcommand\arraystretch{1.5}
	\centering
	\resizebox{\linewidth}{!}
	{
	\begin{tabular}{llcc}
		\hline
		Symbol & Definition \\ \hline 		
		$\mathcal{G}$= $(\mathcal{V},\mathcal{E}, \mathbf{X})$ &Graph with node features\\
		$v_i$  & Node $ i $ \\
		$\mathbf{A} ^{ N\times N}$ & Adjacency matrix for the network \\
		$\mathbf{X} \in \mathbb{R}^{ N\times D}$& Matrix of nodes features \\
		$\vectors{{x}}_i$ & Feature vector for $v_i$\\
		$\vectors{h}_{i}^{(0)}$ & The first-level representation of $v_i$\\ 
		$n$ & fixed node bandwidth via neighbor sampling\\
		$\mathbf{X'}_{i} \in \mathbb{R}^{D \times n}$  & Reconstructed feature map for $v_i$  \\
		$\vectors{h}_{i}^{(K)}$ & The hidden representation of $K$-th GCN layer\\
		$\mathbf{Y}_{lf} \in \mathbb{R}^{\left| \mathcal{V}_l \right|\times F}$ &  Label indicator matrix \\
		\hline
	\end{tabular}
    }
	\caption{List of important notations}
    \label{tab:notation}
    \end{table}

\subsection{Graph Representation Learning}
Traditional machine learning methods on graphs are task-dependent and require feature engineering. In contrast, the more data-driven graph representation learning approach aims to learn task-independent representations that can capture rich information in graphs for various down-stream tasks such as node classification, link prediction, and recommendation. For graphs with associated features as defined above, the graph representation will be learned from both the structure information defined by the nodes $\mathcal{V}$ and edges $\mathcal{E}$, as well as the features $\mathbf{X}$.

\citet{Hamilton2017RepresentationLO} categorizes graph representation leaning methods into three approaches: the factorization-based approach, random walk-based approach and neural network-based approach. 
\begin{enumerate}
	\item \textbf{Factorization-based methods.} Early methods for learning node representations largely focused on matrix factorization. They are directly inspired by classic techniques for dimensionality reduction \cite{belkin2002laplacian, Ahmed2013DistributedLN}.
	
	\item \textbf{Random walk-based methods.} Inspired by the Word2Vec method \cite{mikolov2013distributed}, \citet{perozzi2014deepwalk} proposed the DeepWalk that generates random paths over a graph. It learns the new node representation by maximizing the co-occurrence probability of the neighbors in the walk. Node2vec \cite{grover2016node2vec} and LINE \cite{tang2015line} extend DeepWalk with more sophisticated walks. PLANTOID learns the embedding from both labels and graph structure by injecting the label information \cite{Yang:2016:RSL:3045390.3045396}. 
	
	\item \textbf{Neural network-based methods.} Graph neural networks (GNNs) have previously been introduced by \citet{Gori2005ANM} and \citet{Scarselli2009theGN}, which consist of an iterative process propagating the node states until the node representation reaches a stable fixed point. More recently, several improved methods have been proposed. \citet{Li2015GatedGS} introduced gated recurrent units \cite{Cho2014LearningPR} to alleviate the restriction. \citet{duvenaud2015convolutional} further introduced a convolution-like propagation rule on graphs, which does not scale to large graphs with wide node degree distributions. 
\end{enumerate}
\subsection{Graph Convolutional Networks (GCN)}
The above graph representation methods mainly consider the graph structure (node and edge) information but they do not use the node feature matrix $\mathbf{X}$ in the learning process. \citet{kipf2016semi} proposed the graph convolutional networks (GCN) as an effective graph representation model that can naturally combine structure information and node features in the learning process. It is derived from conducting graph convolution in the spectral domain \cite{bruna2013spectral} \cite{Cho2014LearningPR}. It represents a node by aggregating feature vectors from its neighbors (including itself), which is similar with the convolution operation in CNN. The propagation rule of GCN can be summarized by the following expression:
\begin{equation}
\textstyle
\mathbf{H}^{(l+1)}= \sigma\!\left(\mathbf{\hat{A}H}^{(l)} \mathbf{W}^{(l)} \right) \,,
\label{eq:gcn1layer}
\end{equation}
where
\begin{equation}
\mathbf{\hat{A}} = \mathbf{\tilde{D}}^{-\frac{1}{2}} \mathbf{\tilde{A}}\mathbf{\tilde{D}}^{-\frac{1}{2}}
\end{equation}
is a normalized adjacency matrix of the undirected graph $\mathcal{G}$ with added self-connections
\begin{math}
\mathbf{\tilde{A} = A + I_N} 
\end{math}.
$\mathbf{I_N}$ is an identity matrix. The diagonal entries of $\mathbf{\tilde{D}}$ is
\begin{math}
\mathbf{\tilde{D}}_{ii} = \sum_j \mathbf{\tilde{A}}_{ij}.
\end{math}
$\mathbf{W}^{(l)}$ is a layer-specific trainable weight matrix, $\sigma(\cdot)$ denotes an activation function such as the $\mathrm{ReLU}(\cdot) = \max(0,\cdot)$, and $\mathbf{H}^{(l)}\in \mathbb{R}^{N\times D}$ is the matrix of activation in the $l th$ layer. $\mathbf{H}^{(0)}=\mathbf{X}$ is the node feature matrix.

In Eq. (\ref{eq:gcn1layer}), \begin{math} \mathbf{\hat{A}}_{i}\mathbf{H}^{(l)} \end{math} can be treated as: average with different weights (according to the node degrees) of the central node and all its neighbors' feature vectors.
\begin{equation} 
\mathbf{\hat{A}}_{i}\mathbf{H}^{(l)} = MEAN( \vectors{h}^{(l)}_{i} + \sum_{j \in \mathcal{N}_{i}} \vectors{h}^{(l)}_{j}, j \in \mathcal{N}_{i} ).
\label{eq:gcn_average}
\end{equation}

Here, we can call Eq. (\ref{eq:gcn1layer}) as a GCN layer consisting of  two step: 1) averaging the central node and its neighbors' feature vectors with different weights (according to the node degree), then 2) feeding the averaged feature vector to a fully-connected networks to get a new representation. 

GCN has significantly advanced the state-of-the-art in graph representation learning, particularly in the problem of semi-supervised node classification. However, there are still two major limitations. 
\begin{enumerate}
	\item \textbf{Neighborhood selection/weighting.}
	Equation (\ref{eq:gcn_average}) shows that GCN learns the new node representation from features of all its neighbors, not considering whether the node has a dense or sparse connection. Real-world graphs can have node degrees ranging from one to hundreds or even thousands. Therefore, some nodes may need more neighbors to get sufficient information, while some other nodes may aggregate too broadly such that their own features $\vectors{h}^{(l)}_{i}$ may be ``washed out'' due to aggregating too many $\vectors{h}^{(l)}_{j}$ in Eq. (\ref{eq:gcn_average}). 
	
	\item \textbf{Feature selection/weighting.}
	GCN did not select or weight the features in a feature vector. In this case, noisy features can be aggregated to produce the new representation, which can confuse the classifier and reduce the classification accuracy. 
\end{enumerate}

\subsection{GCN Extensions}
There are two major approaches to deal with the two problems mentioned above:  sampling-based methods and feature convolution-based methods 

\begin{itemize}
	\item \textbf{Sampling-based methods.} These methods aim to get a fixed number of neighbors for each node, to get closer to the CNN application scenario of fixed neighborhood size. GraphSAGE \cite{hamilton2017inductive} uniformly samples a fixed number of neighbors, instead of sampling the full neighbors. These neighbors are generated by a fixed-length random walk, and the neighbors can come from a different number of hops, or search depth, away from a central node. Another sampling-based algorithm is FastGCN \cite{Chen2018FastGCNFL}. It interprets graph convolutions as integral transforms of embedding functions and directly samples the nodes in each layer independently. JP-netwroks \cite{pmlr-v80-xu18c} proposed to sample nodes for the central node from the whole graph rather than the neighbors.
	
	\item \textbf{Feature convolution-based methods.} GCN aggregates the feature vectors of central node and all its neighbors, where each neighbor is treated differently according to their node degree. Inspired by the attention mechanisms \cite{Bahdanau2014NeuralMT}, GAT \cite{velickovic2017graph} learns weights for different neighbors by calculating the correlation between the central node and neighbors' feature vectors via convolution. However, all features in a feature vector share the same weight. The features, which can be useful or not useful, are treated equally without considering their different importance. \citet{Gao:2018:LLG:3219819.3219947} takes GCN features as input and uses two 1D convolutional layers to perform convolution on GCN features for the central node and reorganized features selected from its neighbors. Thus, LGCN inherits the limitations of GCN and the original features are still aggregating over all neighbors without selection due to the GCN layers in front.
\end{itemize}

\subsection{CNN Revisit}
The convolution operation in GCN and its extensions is inspired by that in CNN. Aggregating neighboring node information defined by connectivity is similar to aggregating neighboring pixels in receptive fields for images. While being successful on images, CNN relies on grid-like structures, which is however lacking clear definition in graphs, and filters for convolution operations on the receptive field. In images, most pixels have regular neighbors for the defined receptive field. For graphs, the number of neighbors for each node varies a lot. The \textit{sampling-based} GCN methods addressed this problem by using sampling to get a fixed number of neighbors for each node, which helps defining a fixed-size ``receptive field" for graphs. The \textit{feature convolution-based} method employs 1D convolutional layer on single node features. This motivated our investigation of proposing a convolutional layer on the node-feature maps in graphs. 

\section{The Proposed NFC-GCN}

In this section, we present our proposed approach: graph convolutional networks with node-feature convolution (NFC-GCN). Our method combines ideas from standard GCN, as well as its sampling-based and feature convolution-based extensions, which enables us to design convolution operations on feature maps constructed from feature vectors from the central node and its sampled neighbors. Our operation makes further progress in imitating CNN on graphs. 

\begin{figure}
	\centering
	\includegraphics[width=0.45\textwidth]{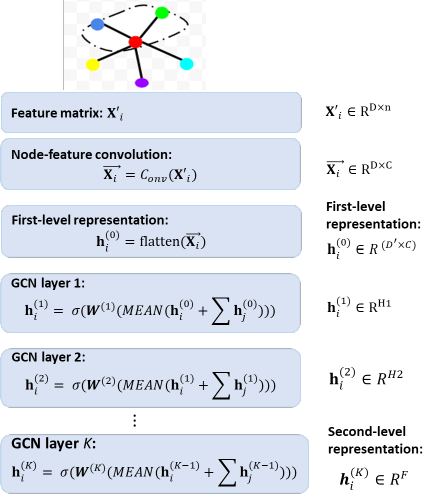}
	\caption{Workflow of NFC-GCN}
	\label{fig:workflow}
\end{figure}

NFC-GCN has three steps: neighbor sampling, node-feature convolution, and standard GCN, as shown in Fig.~\ref{tab:newmodel}. Firstly, neighbors are sampled for each node for a fixed node bandwidth in a graph, which effectively creates a 2D node-feature map that enables the application of a convolutional layer. Then, the convolutional layer works on the 2D node-feature map to get representations of fixed sizes, which are then flattened into a vector as the first-level NFC representation. Next, this NFC representation is fed into a standard GCN to get the second-level NFC-GCN representation, which can be used for downstream tasks such as node classification, link prediction or node recommendation. The workflow is shown as Fig.~\ref{fig:workflow} and Algorithm~\ref{alg:algorithm} gives the pseudocode for one training epoch, with details for each step discussed below.
\subsection{Neighbor Sampling}
For subsequent convolution operations, we need a fixed sized feature map for each node in a graph. Since the node degrees vary greatly across nodes, we use sampling techniques to tackle the problem. For computational simplicity, we use simple random sampling with uniform distribution to select neighbors for each node, although more advanced sampling techniques such as those in \cite{niepert2016learning, hamilton2017inductive, Chen2018FastGCNFL} can be applied in future work. When the number of neighbors is less than the desired size, we duplicate the central node.

\textbf{Feature map.} Let $n$ denote the desired node bandwidth for each node and $d_{i}$ denote the number of the neighbors for node $v_{i}$, which is the degree of $v_{i}$. For each node $v_i$, we will have a local feature map $\mathbf{X'}_{i} \in \mathbb{R}^{D \times n}$, consisting of $n$ $D$-dimensional feature vectors from $n$ nodes consisting of the central node $i$ and its sampled neighbors $\{j'\}$,
\begin{equation}
\mathbf{X'}_{i} = \left\{\vectors{x}_{i}, \vectors{x}_{j'}, j' \in \mathcal{N'}_{i} \right\}_n,
\label{eq:feature map}
\end{equation}

where $\mathcal{N'}_{i}$ represents the selected neighbors of node $i$, $\vectors{x}_{i} \in \mathbb{R}^{D}$ and $\vectors{x}_{j'} \in \mathbb{R}^{D}$ represent the feature vector of the central node $i$ and its neighbor $j'$ respectively. On the whole, this leads to a virtual 2D feature map of size $D \times (n\times N)$ for the whole graph, which resembles a 2D image.

In practice, sparsely connected nodes may have less than $n-1$ neighbors. In this case, we use all existing neighbors and duplicate the central node to reach the desired size of $D \times n$ for the local feature matrix $\mathbf{X'}_{i}$. We will analyze the effect of $n$ on node classification performance in Section 4. 

\subsection{Node-Feature Convolution}
The goal of the second stage is to learn to aggregate neighbors and central node and obtain new node representations that can facilitate the subsequent classification tasks. After obtaining a fixed-size local feature matrix $\mathbf{X'}_{i}$ for each node in the first stage, it is natural to introduce convolutional operations to assign different weights to different features of different neighbors during the aggregation process as shown in Fig.~\ref{tab:newmodel} because the local feature matrix is formed as fixed-size grid-like structure. Specifically, we use 1D convolutional layer on the local feature matrix $\mathbf{X'}_{i} \in \mathbb{R}^{D \times n}$ of each node 
\begin{equation}
\vec{\mathbf{X}}_{i} = C_{onv}(\mathbf{X'}_{i}).
\label{eq:convolution}
\end{equation}

The input channel is set to be $n$. The output $\mathbf{\vec{X}}_{i}$ is shaped as $\in \mathbb{R}^{D' \times C}$, where $D'$ is determined by the  filter size $k$ and stride $s$, and $c$ is the filter number. These three convolutional parameters, $k$, $s$, $c$, are also the hyper-parameters of our approach. Alternatively, we can use 2D convolutional layer on each local feature matrix. In this case, the input channel is set to be 1 and we use 2D convolutional layers with size $k_1 \times k_2$ to slide each feature matrix. After both 1D or 2D convolutional operations, we flatten the output as a vector $\vectors{h}_{i}^{(0)}$ as shown in Eq. (\ref{eq:flatten}) to serve as a new node representation for subsequent classification tasks.
\begin{equation}
\vectors{h}_{i}^{(0)} = flatten(\vec{\mathbf{X}}_{i}).
\label{eq:flatten}
\end{equation}

\subsection{GCN Layers}
The last stage is quite straightforward, in which we directly feed the learned node representation vectors into GCN layers. The GCN layers take the element-wise mean of the vectors
\begin{math} \left\{\vectors{h}_{i}^{(0)}, \vectors{h}_{j}^{(0)}, i \in \mathcal{V}, j \in \mathcal{N}_{i}, \right\} \end{math} to learn the new representation of node $i$, and it can be written as: 
\begin{equation}
\vectors{h}^{(1)}_{i} = \sigma(\mathbf{W}^{(1)}(MEAN(\vectors{h}_{i}^{(0)}+  \sum_{j \in \mathcal{N}_{i} }\vectors{h}_{j}^{(0)}))),
\label{eq:nfc_gcn average}
\end{equation}
where $\vectors{h}^{(1)}_{i}$ is the representation of $v_i$ after the first GCN layers.

After $K$ GCN layers, the final representation $\vectors{h}^{(K)}_{i}$ will be passed to a one-layer neural networks with a $softmax$ activation function. For multi-class classification, the loss function is defined as the cross-entropy error over all labeled examples:
\begin{equation}
\mathcal{L}= -\sum_{l\in\mathcal{V}_l}\sum_{f=1}^F \mathbf{Y}_{lf} \ln \vectors{h}^{(K)}_{l}\, ,
\label{eq:loss}
\end{equation}
where $\mathcal{V}_l$ is the set of node indices that have labels and $F$ is the dimension of output features that is equal to the number of classes. $\mathbf{Y}_{lf} \in \mathbb{R}^ {\left| \mathcal{V}_l \right|\times F} $ is a label indicator matrix.

\begin{table*}
	\centering	
	\begin{tabular}{lccccc}
		\toprule
		Dataset&Nodes&Edges&Features&Classes&Training/Validation/Test \cite{Chen2018FastGCNFL})\\
		\midrule
		Cora & 2708&5429&1433&7&1208/500/1,000\\
		Citeseer &3327&4732&3703&6&1827/500/1,000\\
		PubMed & 19717&44338&500&3&18217/500/1,000\\
		\bottomrule
	\end{tabular}
\caption{Overview of the three datasets}
\label{tab:datasets}
\end{table*} 

\begin{algorithm}[tb]
	\centering
	\caption{NFC-GCN}
	\label{alg:algorithm}
	\begin{algorithmic}
		\STATE {\bfseries Input:} 
		 	   $\mathcal{G}$= $(\mathcal{V},\mathcal{E}, \mathbf{X}$) with $N$ nodes;\\
				Adjacency matrix $\mathbf{A}\in\mathbb{R}^{N\times N}$;\\
				Feature matrix $\mathbf{X} \in \mathbb{R}^{N \times D}$;\\
				Labeled node $\mathcal{V}_l$;\\
				Label indicator matrix $\mathbf{Y}_{lf} \in \mathbb{R}^{\left| \mathcal{V}_l \right|\times F}$;\\
				The fixed-size set of neighbors is ($n$-1);\\
				The parameters in the node-feature convolution process: filter size $k$, stride $s$, the number of filters:$c$, the convolution operation $C_{onv}(\cdot)$
		\STATE {\bfseries Output:} Vector representation  $\vectors{h}_{i}^{(K)}$
		
		\FOR {each $v_{i}$ $\in$ $\mathcal{V}_l$}
				\IF {  $d_{i}$ $>$ $n-1$ }
				\STATE random sample $n-1$ neighbors
				\ELSIF {}
				\STATE duplicate $v_{i}'s$ feature vector $(n-1-d_{i})$ times
				\ENDIF
				\STATE \begin{math}
                \mathbf{X'}_{i} = \left\{\vectors{x}_{i}, \vectors{x}_{j'}, j' \in \mathcal{N'}_{i} \right\}_n
				\end{math}		
				\STATE  $\vec{\mathbf{X}}_{i}$ = $C_{onv}$($\mathbf{X'}_{i}$)
				\STATE 
				\begin{math}
				\vectors{h}_{i}^{(0)} = flatten(\vec{\mathbf{X}}_{i})
				\end{math};
				
				\FOR{each layer $k$, $k$=1,...,$K$}
				\STATE 
			    \begin{math}
				{\vectors{h}^{(k)}_{i} = \sigma(\mathbf{W}^{(k)}(MEAN(\vectors{h}^{(k-1)}_{i}+  \sum_{j \in \mathcal{N}_{i} }\vectors{h}^{(k-1)}_{j})));}
				\end{math}
				\ENDFOR		
				\ENDFOR

	\end{algorithmic}
\end{algorithm}

\subsection{Relationship with Highly Related Work} 
Both, our method and GAT add one layer before GCN. In Eq. (\ref{eq:nfc_gcn average}), the input of GCN model can be treated as $\vectors{h}_{i}^{(0)}=\vectors{{x}}_{i}$ which is the raw feature vector. In GAT model is no longer the raw feature vector, and it is the raw feature vector' embedding containing only its own features.  $\vectors{h}_{i}^{(0)}$ in our method is a more higher-level representation that contains node $i$ and part of its neighbors (local graph structure) information.
Besides, in the GCN stage, the input of our model has been carefully selected by the filters in the node-feature convolution stage, while GCN and GAT consider all the neighbors without any node or feature selection. 
\begin{table*}
	\centering
	\begin{tabular}{lccc}
		\toprule
		\\[-1em]
		Dataset&Cora&Citeseer&PubMed\\
		\midrule
		Input& 2708 $\times$ 1433 $\times$ 6 $\times$ 1 & 3327$\times$ 3703$\times$ 6 $\times$ 1&19717$\times$ 500 $\times$ 6$\times$ 1\\\hline
		& filter = $\left[ 32 \times 6 \right]$ $\times$ 64& filter = $\left[ 64 \times 6 \right]$ $\times$64& filter = $\left[ 64 \times 6 \right]$ $\times32$\\	
		Convolutional layer&stride = 16& stride=32&stride=16\\\hline
		GCN layer 1&16&16&32 \\\hline
		GCN layer 2&7&6&3 \\\hline
		\\[-1em]Classifier layer &7&6&3\\ \hline
	\end{tabular}
\caption{The parameters for our model. For Cora, Citeseer and PubMed, we both choose 5 neighbors in the first convolution operation, then we use 2 GCN layers and 1 classifier layer for the node classification.}
\label{tab:parameter}
\end{table*}
\section{Experiments}

In this section, we have performed evaluation of our models against a wide variety of strong baselines and previous approaches on three citation networks---Cora, Citeseer and PubMed. Then, we list the comparative methods. Finally, we present the experiments results and analyse the advantages and limitations of our method.  

\subsection{Datasets}
We utilize three citation network benchmark datasets---Cora, Citeseer and PubMed \cite{sen2008collective}, with the same train/validation/test splits in \cite{Chen2018FastGCNFL}, as summarized in Table~\ref{tab:datasets}. Detailed descriptions are given below.

\begin{itemize}
	\item \textbf{Cora}\quad The Cora dataset contains 2,708 documents (nodes) classified into 7 classes and 5,429 citation links (edges). We treat the citation links as (undirected) edges and construct a binary, symmetric adjacency matrix. Each document has a 1,433 dimensional sparse bag-of-words feature vector and a class label. 
	\item \textbf{Citeseer}\quad The Citeseer dataset contains 3,327 documents classified into 6 classes and 4,732 links. Each document has a 3,703 dimensional sparse bag-of-words feature vector and a class label.
	\item \textbf{PubMed} \quad The PubMed dataset contains 19,717 documents classified into 3 classes and 44,338 links. Each document has a 500 dimensional sparse bag-of-wordss feature vector and a class label. 
\end{itemize}

\subsection{Experimental Set-up}

We train a three-layers model (one convolutional layers and two GCN layers) and evaluate prediction accuracy on Cora, Citeseer and PubMed datasets. We choose the datasets splits as shown in Table~\ref{tab:datasets}, which is the same as in Fast-GCN \cite{Chen2018FastGCNFL}. We choose a big proportion training datasets, because our model has more parameters to learn than GCN. We use an additional validation set of 500 labeled examples for hyperparameters optimization. Throughout the experiments, we use the Adam optimizer \cite{kingma2014adam} with learning rate 0.002 for Cora and Citeseer, 0.01 for PubMed. We fix the dropout rate to be 0.5 for the hidden layers' inputs and add an L2 regularization of 0.0001. We employ the early stopping strategy based on the validation accuracy and train 200 epochs at most.

We choose 5 neighbors for the central node and the remaining parameters are summarized in Table~\ref{tab:datasets} for our method with 1D convolutional layer. Besides, we also use a 2D convolutional layers in the node-feature convolution stage. We set the width of filter is 3 and all the remaining parameters are the same as the 1D convolutional lays as shown in Table~\ref{tab:parameter}.

\subsection{Baselines}

We compare against four state-of-the-art baselines. In order to ensure the baselines have sufficient diversity, we compare against the following methods:
\begin{itemize}	
	\item \textbf{GCN.} Graph convolutional networks (GCN ) is the most important baseline. In the experiment, we use a two-layer GCN model. The main parameters are set as the original paper: 0.5 (dropout rate), 16 (number of hidden units), 10 (early stopping), 0.1 (learning rate). For there are more training data, we set max training epoch is 400. We use code \footnote{https://github.com/tkipf/gcn} publicly available for GCN.
	
	\item \textbf{Sampling based methods.}  We choose FastGCN and GraphSAGE as the comparison methods. We use the same data splits as FastGCN and we resuse the results of FastGCN and GraphSAGE reported in  \cite{Chen2018FastGCNFL}.
	\item \textbf{Aggregation based.} We choose the most  related one: GAT as the comparison method. It learns to assign different weight to different neighbors. In the experiment, we use the code publicly available for GAT \footnote{https://github.com/PetarV-/GAT}. We use a two-layer GAT model. The main parameters are set as the original paper: 8 (attention heads), 8 (each feature dimension after the 1D convolution). We stop the training within 400 epochs. 
\end{itemize}

\section{Results}
    \begin{table}
	\centering
	\begin{tabular}{lccc}
		\toprule
		Methods&Cora&Citeseer&PubMed\\
		\midrule
		GCN& 86.3\% &77.8\%&86.8\%\\
		Fast-GCN&85.0\% &77.6\%&88.0\%\\
		GraphSAGE-mean& 82.2\% &71.4\%&87.1\%\\
		GAT&80.4\% &75.7\%&85.0\%\\\hline
		NFC-GCN(1D)& \textbf{88.3\%} &\underline{78.5\%}&\textbf{89.5\%}\\
		NFC-GCN(2D)& \underline{88.0\%} &\textbf{78.9\%}&\underline{88.43\%}\\
		\bottomrule
	\end{tabular}
	\caption{Prediction results of node classification on Cora, Citeseer, and PubMed datasets on FastGCN's data splits. (\textbf{Best}; \underline{Second best})}
    \label{tab:results}
\end{table}
\subsection{Semi-supervised Node Classification}
\begin{itemize}	[leftmargin=*]

	\item 	\textbf{Test accuracy.} The results of our comparative evaluation experiments are summarized in \textbf{Table~\ref{tab:results}}. We report the classification accuracy (average of ten runs) on the test nodes of our methods, and reuse results already reported in Fast-GCN \cite{Chen2018FastGCNFL}. We run GCN and GAT in the same data splits setting and it is curious that GCN outperforms GAT in our dataset split, though consistent with the experiments results in \cite{pmlr-v80-xu18c}. FastGCN and GraphSAGE-mean can not perform as well as GCN, because they only use limited nodes in the graph.

    \begin{figure}[h]
    	\centering
    	\begin{subfigure}[b]{0.43\textwidth}
    		\includegraphics[width=\textwidth]{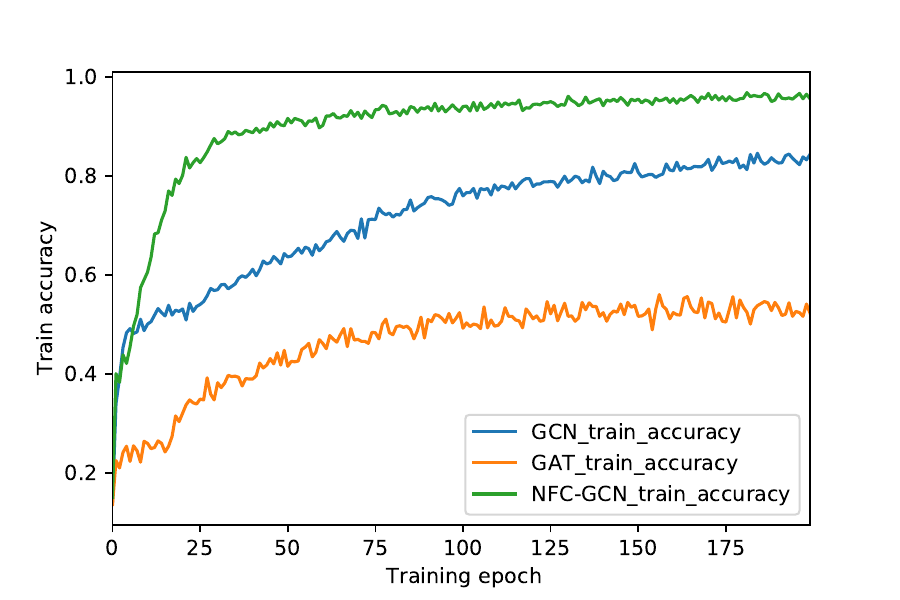}
    		\caption{Cora training loss changes with training epoch}
    	\end{subfigure}
   
    	\begin{subfigure}[b]{0.43\textwidth}
    		\includegraphics[width=\textwidth]{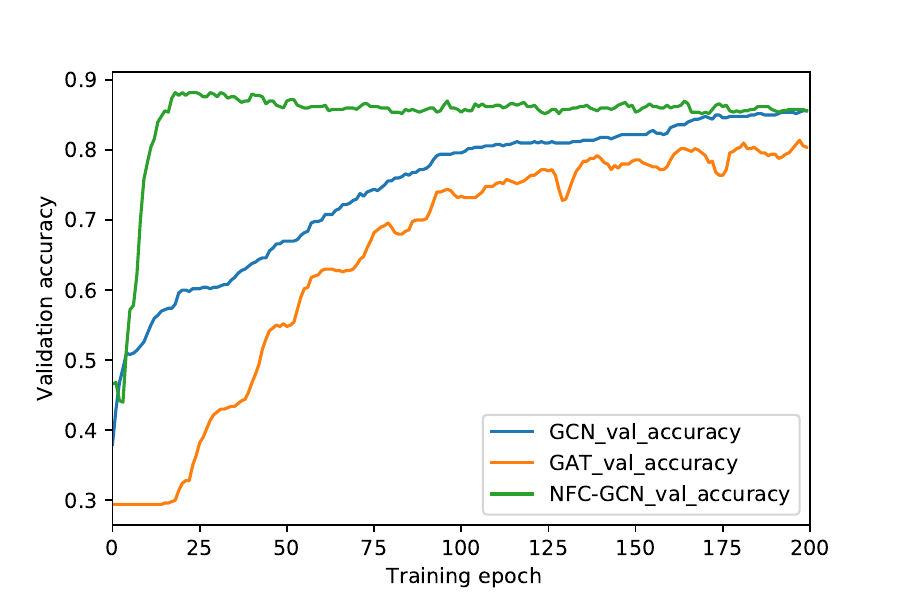}
    		\caption{Cora validation loss changes with training epoch}
    	\end{subfigure}
   
    	\caption{In this experiment, we train the models for 200 epoch (without early stopping). It can be seen that our method can quickly get better results in less than 50 training epochs}\label{fig:accuracy}
    \end{figure}

 \begin{figure}[h]
	\centering
	\begin{subfigure}[b]{0.45\textwidth}
		\includegraphics[width=\textwidth]{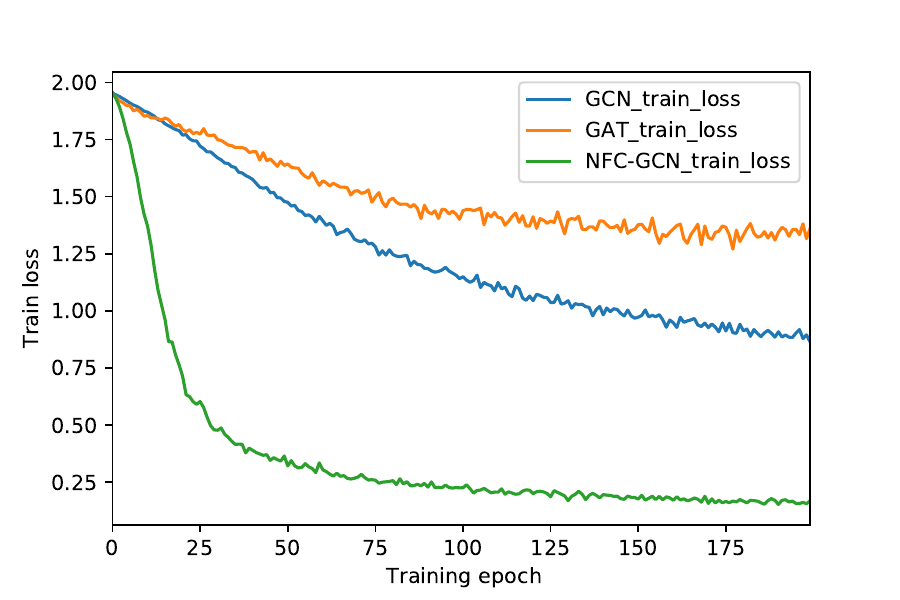}
		\caption{Cora training loss changes with training epoch}
	\end{subfigure}
	
	\begin{subfigure}[b]{0.45\textwidth}
		\includegraphics[width=\textwidth]{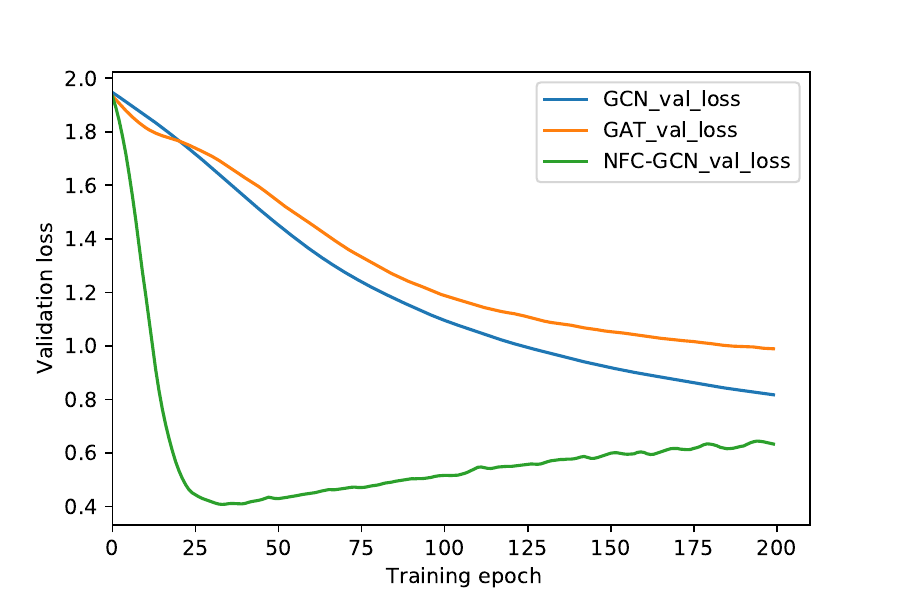}
		\caption{Cora validation loss changes with training epoch}
	\end{subfigure}
	
	\caption{In this experiment, we train the models for 200 epoch (without early stopping). It can be seen that after the node-feature convolution process, the new node representation can remarkably facilitate the subsequent classification tasks.}\label{fig:loss}
\end{figure}
    Our results successfully demonstrate state-of-the-art performance across all datasets, even we also only use limited nodes as FastGCN and GraphSAGE. Besides, we try to use 1D convolutional layer and 2D convolutional layer in the node-feature convolution separately. They both outperform other methods on all the three datasets, and in the experiments 1D convolutional layer performs better than 2D convolutional layer on Cora and PubMed. We are able to improve upon all the methods by a margin of \textbf{2.0\%} on Cora, suggesting that learning to wisely aggregate the fixed size neighbors by using NFC layer can be greatly beneficial.
	In addition, our method can get the better performance in a small of training epochs, but our training time is much more than other methods, because our model is much more complex and there are many parameters to learn in each training epoch.  
    \begin{table}
	\centering
	\begin{tabular}{cccl}
		\toprule
		Methods&Cora&Citeseer&PubMed\\
		\midrule
		GCN\_5& 64.8\% &74.1\%&80.0\%\\
		GAT\_5&64.2\% &74.2\%&82.2\%\\
		NFC\_5-GCN & \textbf{86.0\%} &\textbf{79.1\%}&\textbf{89.0\%}\\\hline
		Improvement& 21.2\%&4.9\%&6.8\%\\
		\bottomrule
	\end{tabular}
\caption{Results of test accuracy without the GCN layer.
	GCN\_5 means aggregate the central and five neighbors' features to get the new representation of central node in GCN's manner. GAT\_5 means average the central and five neighbors' features with learned weights to get the new representation of central node. NFC\_5-GCN uses the convolutional layer to learn the new representation of central node and five neighbors' features to get the new representation. Then, feeding the new representation to a one layer neural network to get the classification results.}
\label{tab:0layers}
\end{table}
	\item \textbf{Accuracy, loss change with training epoch.} Here, we also show the training accuracy, validation accuracy, training loss, and validation loss change with each training epoch in \textbf{Fig.~\ref{fig:accuracy}, Fig.~\ref{fig:loss}.} We did not use early stopping in our model for a better comparison with GCN and GAT in the same training epoch and we show the results within 200 epochs. We can see that our method can get good results in few training epochs, while GCN and GAT need one hundred training epochs and even more to stable. Besides, we can see that training, validation accuracy/loss of NFC-GCN rise or descend not only very quickly but also stably. With the same Adam SGD optimizer to minimize cross-entropy on the training nodes as GCN and GAT, our method performs better in the optimization process. This verifies that the first-level node representation learned from the node-feature convolution can improve the subsequent classification tasks.

	\item \textbf{Test accuracy without GCN layer.} In order to show our methods can learn more smartly from the neighbors features, we use three different ways to deal with a fixed size set of neighbors: average with different weights according to neighbors' node degrees (GCN), learn to assign weight to neighbors and then mean average (GAT), convolution of node and features (ours). Then we feed the new representation of each node to the classifier layer directly. We test this comparison experiments on Cora, Citeseer and PubMed, and each node has 5 neighbors for the central node to use the 3 different ways to learn the new representation for node classification. The results are summarized in \textbf{Table~\ref{tab:0layers}}. In this experiment, we can see that how to deal with the neighbors' features has a significant effect on the final results. Our methods significantly increase the test accuracy by margins of \textbf{21.2\%} on Cora, \textbf{4.9\%} on Citeseer, \textbf{6.8\%} on PubMed. 
	
	\textbf{It should be emphasized that our method can get competitive performance without the GCN layers.} From Table~\ref{tab:results} and Table~\ref{tab:0layers}, we can see that the best performance of other methods for Cora, Citeseer and PubMed are 86.3\%, 77.8\%, 88.0\% respectively, while ours results are 86.0\%, 79.1\%, 89.0\%. Adding the GCN layer can slightly improve the performance (enhancement of about 1$\%$).
\end{itemize}

\subsection{Effect of the Node Bandwidth}

\begin{table}
	\centering
	\begin{tabular}{ccccl}
		\toprule
		Dataset & Highest & Lowest & Average&Median\\ \hline
		Cora & 168  & 1 & 4.87&4  \\                   
		Citeseer& 99& 1 & 3.7&3   \\                  
		Pumbed & 171 & 1& 5.5&3  \\             
		\bottomrule
	\end{tabular}
	\caption{Node degree statistics on Cora, Citeseer and PubMed.}
	\label{tab:nodedegree}
\end{table}

\begin{figure*}
	\centering
	\begin{subfigure}[b]{0.3\textwidth}
		\includegraphics[width=\textwidth]{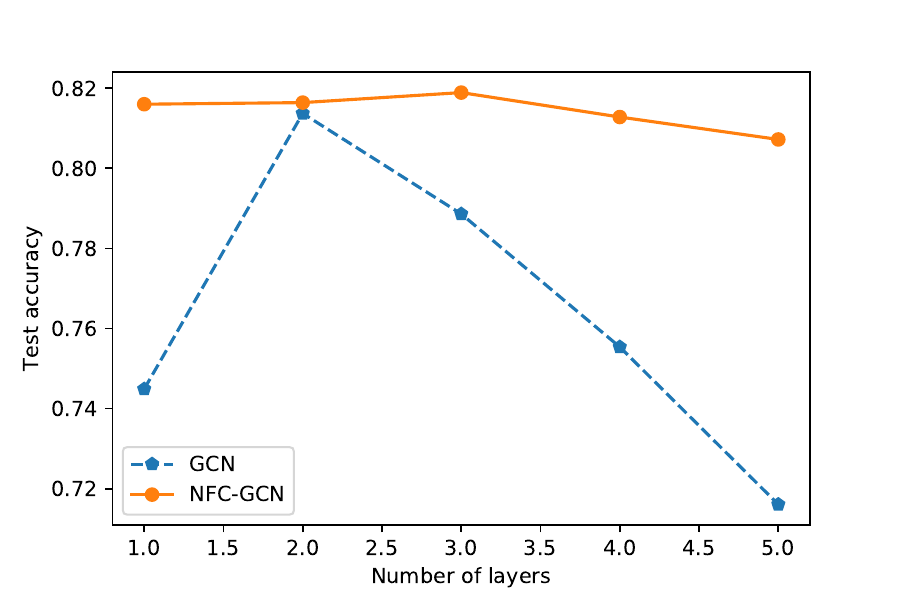}
		\caption{Cora}
	\end{subfigure}
	~ 
	\begin{subfigure}[b]{0.3\textwidth}
		\includegraphics[width=\textwidth]{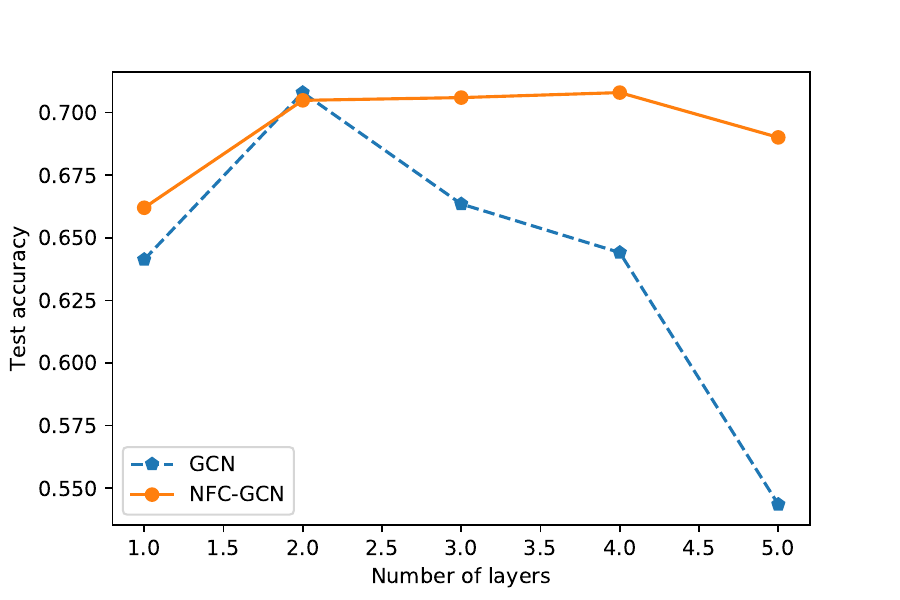}
		\caption{Citeseer}
	\end{subfigure}
	~ 
	\begin{subfigure}[b]{0.3\textwidth}
		\includegraphics[width=\textwidth]{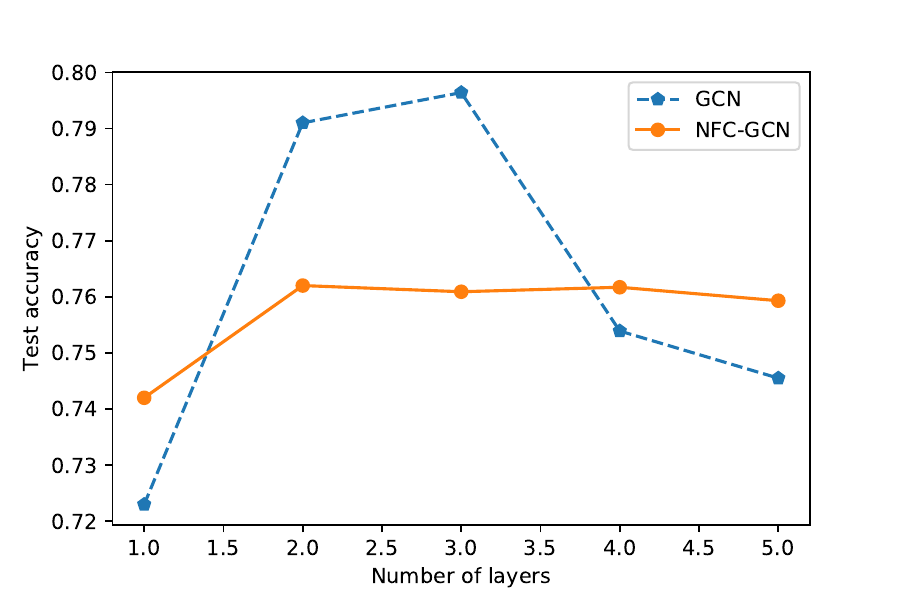}
		\caption{PubMed}
	\end{subfigure}
	\caption{Results of using different GCN layers in NFC-GCN and GCN models. On the Cora, Citeseer and PubMed datasets, we employ the same experimental setups (each class has 20 labeled data for training), we just change the GCN's layers and report the classification accuracy in this figure. It can be seen that GCN-NFC allows for a deeper model, for the test accuracy of our method can keep more steady than GCN with the changing of layers.}\label{fig:change layers to see test accuracy}
\end{figure*}

\begin{table}
	\centering
	\begin{tabular}{cccl}
		\toprule
		Node bandwidth $n$&Cora&Citeseer&PubMed\\
		\midrule
		NFC$_{n=2}$-GCN &87.53\%&78.30\%&87.58\%\\
		NFC$_{n=3}$-GCN &87.79\%&78.34\%&87.83\%\\
		NFC$_{n=4}$-GCN &88.13\%&78.37\%&88.02\%\\ 
		NFC$_{n=5}$-GCN &88.18\%&78.48\%&88.29\%\\
		NFC$_{n=6}$-GCN &88.30\%&78.52\%&88.23\%\\
		\bottomrule
	\end{tabular}
	\caption{Results of varying $n$ in the first-level representation learning. In this experiment, we fix the model architecture: one convolutional layer and 2 GCN layers, and just change $n$ from 2 to 6 in the first convolution operation. NFC$_{n=k}$-GCN denotes a node bandwidth of $n$. It can be seen that test accuracy tends to yield consistent improvement in accuracy with increasing $n$ on the whole.}
	\label{tab:test accuracy changes with the number of neighors}
\end{table}
In the node-feature convolution process, we fix the node bandwidth $n$ to get a fixed size feature map for the central node and enable the following convolution operation. In order to see the influence of the node bandwidth, we study the effect of varying $n$ in the node-feature convolution process. 

Table~\ref{tab:nodedegree} shows the distribution of node degrees $d_i$ for the three datasets studied. We can see that the three datasets are very sparse graphs, so we vary $n$ from 2 to 6. For a larger $n$, e.g., $n=6$, there are many duplication of the central nodes. and the respective results are summarized in the \textbf{Table~\ref{tab:test accuracy changes with the number of neighors}}. The results show consistent improvement in accuracy with increasing $n$. A larger $n$ means more feature diversity, and this can be especially helpful for the representation learning of nodes with sparse features. However, a larger $n$ will increase the computation cost so in practice, there is a trade-off between the classification accuracy and computation complexity when choosing $n$.

\subsection{Effect of the Model Depth}

Next, we investigate the influence of model depth (number of GCN layers) on classification performance. We change the GCN layers from 1 to 5 and the results are summarized in \textbf{Tabel~\ref{tab:layers}}. We can see that the three citation datasets' test accuracy changes less than $3.50\%$.

Besides, we also notice that the best test accuracy for our methods is not much better than GCN in Cora and Citeseer datasets. The best test accuracy is not better than GCN in PubMed dataset (with only 60 labeled training data), because our model has more parameters to learn, NFC-GCN is more suitable for datasets with many labeled training data. This is one of our method's limitation.

Our method is less sensitive with the number of hidden layers, and this indicates that the new representation of the central node becoming more robust and easily classified after the node-feature convolution process. This can be verified in Table~\ref{tab:0layers}. After the first convolutional process, the new representation of the central node is feed to a one layer neural networks directly and can get much better results than GCN and GAT. This shows that the learned new features are significantly representative for this class. So, even $K$-th order neighborhood is treated as the context size for the central node, and it can be still accurately classified. Besides, we just use limited nodes in our methods, and this can prevent the neighborhood explosion to a certain degree. 
\begin{table}
	\begin{tabular}{cccl}
		\toprule
		NFC-GCN&Cora&Citeseer&PubMed\\
		\midrule
		NFC-GCN$_{1}$&86.58\%&76.70\%&89.49\%\\
		NFC-GCN$_{2}$&88.30\%&78.52\%&88.23\%\\
		NFC-GCN$_{3}$&88.15\%&77.95\%&86.70\%\\
		NFC-GCN$_{4}$ &87.51\%&77.43\%&86.28\%\\
		NFC-GCN$_{5}$&86.83\%&76.88\%&85.77\%\\ \hline	
		Test accuracy changes &1.72\%&1.82\%&3.21\%\\
		\bottomrule
	\end{tabular}
\caption{Effect of the GCN layers' influence. In this experiment, we choose $n=6$ as the bandwidth in first convolution operation and just change the number of GCN layers from 1 to 5 to get the classification accuracy. NFC-GCN$_{k}$ means using $k$ GCN layers.}
\label{tab:layers}
\end{table}

\section{Discussion}
NFC-GCN embodies the ideas from GCN and its extensions such as sampling-based GraphSAGE and feature convolution-based GAT. NFC-GCN simply adds a convolution layer before GCN layers that convolves local feature maps for each node, which differs from the 1D convolution layer in GAT. In the following, we discuss the limitation and new opportunities for this new architecture 

\begin{itemize}	
	\item \textbf{Limitations.} Our method's main limitation is that NFC-GCN needs more training samples to learn the parameters. Although in each training epoch, NFC-GCN has a larger gradient descent than GCN/GAT, the computation cost for each epoch of NFC-GCN is higher than that of GCN. Nonetheless, deep learning models are known to need many samples to work well and be computationally expensive.
	\item \textbf{Powerful representation learning ability.} We have the choice of either 1D or 2D convolution to learn nodes' new representation as $\vec{x}_{i}$, and feeding $\vec{x}_{i}$ directly to a classifier layer can get a competitive performance compared with GCN and its extensions. One important future direction is to totally get rid of the GCN layer to explore an architecture in a more CNN manner. This will enable many ideas and tricks in CNN to be applied to graphs.
	\item \textbf{Good optimization performance.} In the training process, our method can get better performance within 50 training epochs, while GCN and GAT need about 100 (twice) training epochs as shown in Fig.~\ref{fig:accuracy} and Fig.~\ref{fig:loss}. Besides, NFC-GCN's accuracy and loss' curves are more steady than GCN and GAT, which is preferred in neural network training. We will further work on more efficient computation for each epoch and perform a deeper analysis of the optimization process of NFC-GCN to understand its convergence / optimization behaviours. 
	\item \textbf{New architectures and deeper models.} The NFC layer proposed in NFC-GCN can be a  layer or stacked into multiple layers for GCN and its extensions. 
	In particular, NFC-GCN allows for a deeper model, and has the potential to get better classification accuracy with tricks in CNN or better tuning of hyperparameters. Furthermore, deeper models can enable other powerful machine learning technique to be better applied to graphs, such as transfer learning.
\end{itemize}
\section{Conclusions}
In this paper, we proposed a novel neural network model for graphs: graph convolutional networks with node-feature convolution (NFC-GCN). The key idea is to construct fixed size 2D feature maps to enable convolution as in the popular CNN model. We constructed such fixed size feature maps via sampling technique. In the proposed node-feature convolution layer, we used multiple filters to perform convolution on feature maps built from feature vectors of the central node and and its neighbors, which produces the first-level node representation. Then we fed the first-level node representation to a standard GCN model to learn the second-level / final representation suitable for downstream tasks. The filter weights were learned in an \textit{end-to-end} fashion with the whole NFC-GCN such that the model learned to assign adaptive weights to different features of different (central or neighbor) nodes. 

Experimental results showed that the proposed NFC-GCN outperformed competing GCN methods (including both sampling-based and feature convolution-based models) on three different popular citation graphs for node classification. Even without the GCN layer, the first-level NFC representation achieved decent performance. Furthermore, the NFC-GCN model took much fewer epochs to converge compared to GCN and GAT and deeper models based on NFC-GCN shows much less performance variation compared to GCN. On the whole, the proposed NFC-GCN architecture opens many new doors for exploring and advancing representation learning for graphs.

\nocite{langley00}

\bibliography{example_paper}
\bibliographystyle{icml2018}

\end{document}